# Experiments with Random Projection

Sanjoy Dasgupta*
AT&T Labs – Research

## Abstract

Recent theoretical work has identified random projection as a promising dimensionality reduction technique for learning mixtures of Gaussians. Here we summarize these results and illustrate them by a wide variety of experiments on synthetic and real data.

## 1 Introduction

It has recently been suggested that the learning of high-dimensional mixtures of Gaussians might be facilitated by first projecting the data into a randomly chosen subspace of low dimension (Dasgupta, 1999). In this paper we present a comprehensive series of experiments intended to precisely illustrate the benefits of this technique.

There are two main theoretical results about random projection. The first is that data from a mixture of $k$ Gaussians can be projected into just $O(\log k)$ dimensions while still retaining the approximate level of separation between the clusters. This projected dimension is independent of the number of data points and of their original dimension. In the experiments we perform, a value of $10 \ln k$ works nicely. Second, even if the original clusters are highly *eccentric* (that is, far from spherical), random projection will make them more spherical. This effect is of major importance because raw high-dimensional data can be expected to form very eccentric clusters, owing, for instance, to different units of measurement for different attributes. Clusters of high eccentricity present an algorithmic challenge. For example, they are problematic for the EM algorithm because special pains must be taken to ensure that intermediate covariance matrices do not become singular, or close to singular. Often this is accomplished by imposing special restrictions on the matrices.

These two benefits have made random projection the key ingredient in the first polynomial-time, provably correct (in a PAC-like sense) algorithm for learning mixtures of Gaussians (Dasgupta, 1999). Random projection can also easily be used in conjunction with EM. To test this combination, we have performed experiments on synthetic data from a variety of Gaussian mixtures. In these, EM with random projection is seen to consistently yield models of quality (log-likelihood on a test set) comparable to or better than that of models found by regular EM. And the reduction in dimension saves a lot of time.

Finally, we have used random projection to construct a classifier for handwritten digits, from a canonical USPS data set in which each digit is represented as a vector in $\mathbb{R}^{256}$. We projected the training data randomly into $\mathbb{R}^{40}$, and were able to fit a mixture of fifty Gaussians (five per digit) to this data quickly and easily, without any tweaking or covariance restrictions. The details of the experiment directly corroborated our theoretical results.

Another, very popular, technique for dimensionality reduction is principal component analysis (PCA). Throughout this paper, both in conceptual discussions and empirical studies, we will contrast PCA with random projection in order to get a better feel for each.

## 2 High-dimensional Gaussians

### 2.1 Some counter-intuitive effects

An $n$-dimensional Gaussian $N(\mu, \Sigma)$ has density function

$$p(x) = \frac{1}{(2\pi)^{n/2}|\Sigma|^{1/2}} \exp\left(-\frac{1}{2}(x-\mu)^T \Sigma^{-1}(x-\mu)\right).$$

If $\Sigma$ is a multiple of the identity matrix, then the Gaussian is called *spherical*. Some important intuition about the behavior of Gaussians in high dimension can quickly be gained by examining the spherical Gaussian $N(0, \sigma^2 I_n)$. Although its density is highest at the origin, it turns out that for large $n$ most of the probability mass lies far away from this center. A point $X \in \mathbb{R}^n$ chosen randomly from this Gaussian has coordinates $X_i$ which are i.i.d. $N(0, \sigma^2)$. Therefore its expected squared Euclidean norm is $\mathbf{E}(\|X\|^2) = \sum_i \mathbf{E} X_i^2 = n\sigma^2$. In fact, it can be shown quite routinely, by writing out the moment-generating function of $\|X\|^2$, that the distribution of $\|X\|^2$ will be tightly concentrated around its expected value. Specifically,

$$\mathbf{P}(|\|X\|^2 - \sigma^2 n| > \epsilon \sigma^2 n) \leq 2 e^{-n\epsilon^2/24}.$$

---

*Work done while at University of California, Berkeley.

That is to say, for big enough $n$, almost the entire distribution lies in a thin shell of radius approximately $\sigma\sqrt{n}$. Thus the natural scale of this Gaussian is in units of $\sigma\sqrt{n}$.

This effect might arouse some initial skepticism because it is not observable in one or two dimensions. But it can perhaps be made more plausible by the following explanation. The Gaussian $N(0, I_n)$ assigns density proportional to $e^{-\rho^2 n/2}$ to points on the surface of the sphere centered at the origin and of radius $\rho\sqrt{n}, \rho \leq 1$. But the surface area of this sphere is proportional to $(\rho\sqrt{n})^{n-1}$. For large $n$, as $\rho \uparrow 1$, this surface area is growing much faster than the density is decaying, and thus most of the probability mass lies at distance about $\sqrt{n}$ from the origin (Bishop, 1995, exercise 1.4). Figure 1 is a graphical depiction of this effect for various values of $n$.

The more general Gaussian $N(0, \Sigma)$ has ellipsoidal contours of equal density. Each such ellipsoid is of the form $\{x : x^T \Sigma^{-1} x = r^2\}$, corresponding to points at a fixed *Mahalanobis distance* $\|x\|_\Sigma = \sqrt{x^T \Sigma^{-1} x}$ from the center of the Gaussian. The principal axes of these ellipsoids are given by the eigenvectors of $\Sigma$. The radius along a particular axis is proportional to the square root of the corresponding eigenvalue. Denote the eigenvalues by $\lambda_1 \leq \cdots \leq \lambda_n$. We will measure how non-spherical a Gaussian is by its *eccentricity*, namely $\sqrt{\lambda_n/\lambda_1}$. As in the spherical case, for large $n$ the distribution of $N(0, \Sigma)$ will be concentrated around an ellipsoidal shell $\|x\|_\Sigma^2 \approx n$. Yet, if $\Sigma$ has bounded eccentricity, this distribution will also be concentrated, perhaps less tightly, around a spherical shell $\|x\|^2 \approx \lambda_1 + \cdots + \lambda_n = \text{trace}(\Sigma)$.

### 2.2 Formalizing separation

It is reasonable to imagine, and is borne out by experience with techniques like EM (Duda and Hart, 1973; Redner and Walker, 1984), that a mixture of Gaussians is easiest to learn when the Gaussians do not overlap too much. Our discussion of $N(\mu, \sigma^2 I_n)$ suggests that it is natural to define the *radius* of this Gaussian as $\sigma\sqrt{n}$, which leads to the following

**Definition** Two Gaussians $N(\mu_1, \sigma^2 I_n)$ and $N(\mu_2, \sigma^2 I_n)$ are *c-separated* if $\|\mu_1 - \mu_2\| \geq c\sigma\sqrt{n}$. More generally, Gaussians $N(\mu_1, \Sigma_1)$ and $N(\mu_2, \Sigma_2)$ in $\mathbb{R}^n$ are *c*-separated if

$$\|\mu_1 - \mu_2\| \geq c\sqrt{\max\{\text{trace}(\Sigma_1), \text{trace}(\Sigma_2)\}}.$$

A mixture of Gaussians is $c$-separated if its component Gaussians are pairwise $c$-separated.

The intention is that two Gaussians are $c$-separated if their centers are $c$ radii apart. Our choice of *radius* for non-spherical Gaussians $N(\mu, \Sigma)$ is motivated by the observation that points $X$ from such Gaussians have $\mathbf{E}\|X - \mu\|^2 = \text{trace}(\Sigma)$.

In high dimension, a 2-separated mixture corresponds roughly to almost completely separated Gaussian clusters, whereas a mixture that is 1- or $\frac{1}{2}$-separated has slightly more (though still negligible) overlap. What kind of separation should be expected of real data sets? This will vary from case to case. As an example, we did some simple analysis of a collection of 9,709 handwritten digits from USPS, where each digit was represented as a vector in 256-dimensional space. We fit a mixture of ten Gaussians to the data, by doing each digit separately, and found that this mixture was 0.63-separated.

One way to think about high-dimensional $c$-separated mixtures is to imagine that their projections to any one coordinate are $c$-separated. For instance, suppose that measurements are made on a population consisting of two kinds of fish. Various attributes, such as length and weight, are recorded. Suppose also that restricting attention to any one attribute gives a 1-separated mixture of two Gaussians in $\mathbb{R}^1$, which is unimodal and therefore potentially difficult to learn. However, if several (say ten) independent attributes are considered together, then the mixture in $\mathbb{R}^{10}$ will remain 1-separated but will no longer have a unimodal distribution. It is precisely to achieve such an effect that multiple attributes are used. This improvement in terms of better-defined clusters is bought at the price of an increase in dimensionality. It is then up to the learning algorithm to effectively exploit this tradeoff.

It is worth clarifying that our particular notion of separation corresponds to the expectation that at least some fraction of the attributes will provide a little bit of discriminative information between the clusters. As an example of when this is *not* the case, consider two spherical Gaussians $N(\mu_1, I_n)$ and $N(\mu_2, I_n)$ in some very high dimension $n$, and suppose that only one of the attributes is at all useful. In other words, $\mu_1$ and $\mu_2$ are identical on every coordinate save one. We will consider these clusters to be poorly separated – their separation is $O(n^{-1/2})$ – even though clustering might information-theoretically be possible.

## 3 Dimensionality reduction

Dimensionality reduction has been the subject of keen study for the past few decades, and instead of trying to summarize this work we will focus upon two popular contemporary techniques: principal component analysis (PCA) and random projection. They are both designed for data with Euclidean ($L_2$) interpoint distances and are both achieved via linear mappings.

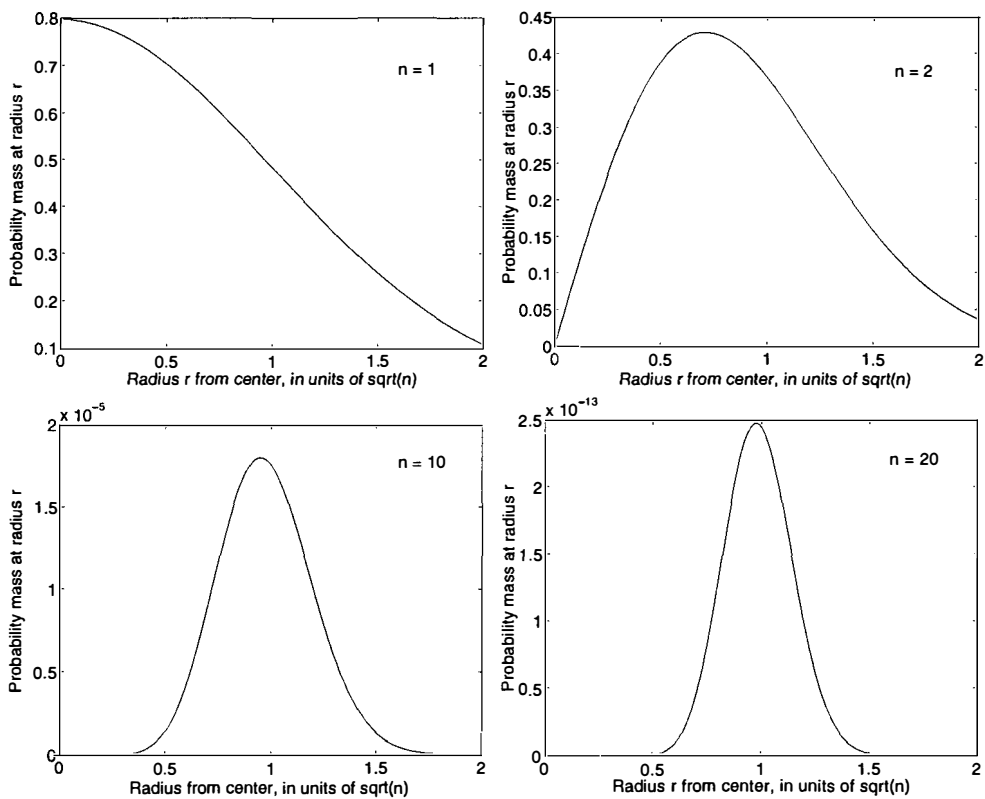

Figure 1: How much of the probability mass of $N(0, I_n)$ lies at radius $r\sqrt{n}$ from the origin? Graphs for $n = 1, 2, 10, 20$.

The linear projection of a Gaussian remains a Gaussian. Therefore, projecting a mixture of high-dimensional Gaussians onto a single line will produce a mixture of univariate Gaussians. However, these projected clusters might be so close together as to be indistinguishable. The main question then is, how much can the dimension be reduced while still maintaining a reasonable amount of separation between different clusters?

## 3.1 Principal component analysis

Principal component analysis is an extremely important tool for data analysis which has found use in many experimental and theoretical studies. It finds a $d$-dimensional subspace of $\mathbb{R}^n$ which captures as much of the variation in the data set as possible. Specifically, given data $S = \{x_1, \ldots, x_m\}$ it finds the linear projection to $\mathbb{R}^d$ for which

$$\sum_{i=1}^{m} \|x_i^* - \mu^*\|^2$$

is maximized, where $x_i^*$ is the projection of point $x_i$ and $\mu^*$ is the mean of the projected data.

PCA is sometimes viewed as one of many possible procedures for *projection pursuit* (Huber, 1985), that is, for finding interesting projections of high-dimensional data. Different notions of "interesting" lead to different projection algorithms. Part of PCA's appeal lies in the fact that it corresponds to an optimization problem which can be solved exactly and efficiently, via eigenvalue computations. Its running time is polynomial, but is nonetheless rather high: $O(n^3)$ for $n$-dimensional data.

By how much can PCA reduce the dimension of a mixture of $k$ Gaussians? It is quite easy to symmetrically arrange a group of $k$ spherical Gaussians in $\mathbb{R}^{k/2}$ so that a PCA projection to any smaller dimension will collapse some of the Gaussians together, and thereby decisively derail any hope of learning. For instance, place the centers of the $(2j - 1)^{st}$ and $2j^{th}$ Gaussians along the $j^{th}$ coordinate axis, at positions $j$ and $-j$. The eigenvectors found by PCA will roughly be coordinate axes, and the discarding of any eigenvector will collapse together the corresponding pair of Gaussians. Thus PCA cannot in general be expected to reduce the dimension of a mixture of $k$ Gaussians to below $\Omega(k)$.

We will next consider a technique which is much faster than PCA, just linear in the dimension, because the choice of projection does not depend upon the data at all. A low-dimensional subspace is picked at random, and it can be shown that with high probability over the choice of subspace, the projected clusters (in that subspace) will retain the approximate level of separation of their high-dimensional counterparts. This use of randomness might seem suboptimal; one might be tempted to assert that "it must be possible to do better by actually taking the data into account". However, we are unaware of any deterministic procedures which have the performance guarantees that will shortly be presented. Randomness is now a standard tool in algorithm design; for instance, it is the basis of the only known polynomial-time algorithm for primality testing.

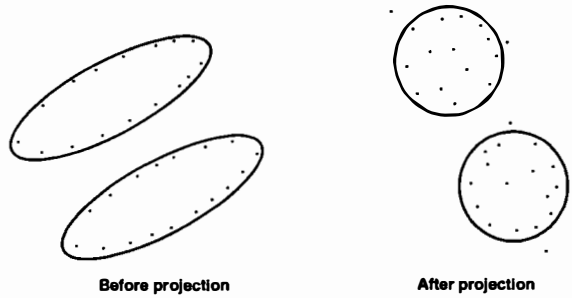

Figure 2: The effects of random projection: the dimension is drastically reduced while the clusters remain well-separated and become more spherical.

### 3.2 Random projection

The following dimensionality reduction lemma applies to arbitrary mixtures of Gaussians. Its statement refers to the notion of separation introduced earlier, and implies that by random projection, data from a mixture of $k$ Gaussians can be mapped into a subspace of dimension just $O(\log k)$.

**Lemma 1 (Dasgupta, 1999)** *For any $c > 0$, pick a $c$-separated mixture of $k$ Gaussians in $\mathbb{R}^n$. Let $\delta, \epsilon \in (0, 1)$ designate confidence and accuracy parameters, respectively. Suppose the mixture is projected into a randomly chosen subspace of dimension $d \geq \frac{C_1}{\epsilon^2} \ln \frac{k}{\delta}$, where $C_1$ is some universal constant. Then, with probability $> 1 - \delta$ over the choice of subspace, the projected mixture in $\mathbb{R}^d$ will be $(c\sqrt{1-\epsilon})$-separated.*

This method of projection has another tremendous benefit: we show that even if the original Gaussians are highly skewed (have ellipsoidal contours of high eccentricity), their projected counterparts will be more spherical (Figure 2). Since it is conceptually much easier to design algorithms for spherical clusters than ellipsoidal ones, this feature of random projection can be expected to simplify the learning of the projected mixture.

**Definition** For a positive definite matrix $\Sigma$, let $\lambda_{max}(\Sigma)$ and $\lambda_{min}(\Sigma)$ refer to its largest and smallest eigenvalues, respectively, and denote by $E(\Sigma)$ the *eccentricity* of the matrix, that is, $\sqrt{\lambda_{max}(\Sigma)/\lambda_{min}(\Sigma)}$.

**Lemma 2 (Dasgupta, 1999)** *Consider any Gaussian in $\mathbb{R}^n$; let $E$ denote its eccentricity. Suppose this Gaussian is projected into a randomly chosen subspace of dimension $d$. There is a universal constant $C_2$ such that for any $\delta, \epsilon \in (0, 1)$, if the original dimension satisfies $n > C_2 \cdot \frac{E^2}{\epsilon^2}(\log \frac{1}{\delta} + d \log \frac{d}{\epsilon})$, then with probability $> 1 - \delta$ over the choice of random projection, the eccentricity of the projected covariance matrix will be at most $1 + \epsilon$. In particular, if the high-dimensional eccentricity $E$ is at most $n^{1/2} C_2^{-1/2}(\log \frac{1}{\delta} + d \log d)^{-1/2}$ then with probability at least $1 - \delta$, the projected Gaussians will have eccentricity at most two.*

Random projection offers many clear benefits over principal component analysis. As explained earlier, PCA cannot in general be used to reduce the dimension of a mixture of $k$ Gaussians to below $\Omega(k)$, whereas random projection can reduce the dimension to just $O(\log k)$. Moreover, PCA may not reduce the eccentricity of Gaussians. These two factors ruled out the use of PCA in the design of a polynomial-time, provably correct (PAC-style) algorithm for mixtures of Gaussians (Dasgupta, 1999). However, if the projected dimension is high enough, then a PCA-projected mixture could easily be far better separated than its randomly projected counterpart. For this reason PCA remains an important tool in the study of Gaussian clusters.

There is a related distinction between PCA and random projection which might be of some practical relevance. As the projected dimension is decreased, as it drops below $\log k$, random projection suffers a gradual degradation in performance. That is, the separation between the clusters is slowly eroded. On the other hand, the degradation suffered by PCA is not necessarily so gradual. In the "symmetric" example considered earlier, any PCA projection to $d \geq k/2$ dimensions has good separation, while a projection to $k/2 - 1$ dimensions has zero separation. Admittedly, this is in part an artefact of our particular definition of separation.

A random projection from $n$ dimensions to $d$ dimensions is represented by a $d \times n$ matrix. It does not depend on the data and can be chosen rapidly. Here is an algorithm which generates the correct distribution over matrices.

- Set each entry of the matrix to an i.i.d. $N(0, 1)$ value.

- Make the $d$ rows of the matrix orthogonal by using the Gram-Schmidt algorithm, and then normalize them to unit length.

This takes $O(d^2 n)$ time overall. An even faster method, which takes time only $O(dn)$, is simply to choose each entry of the matrix uniformly and independently from $[-1, 1]$. This does not precisely generate the distribution we want but will also work well (Achlioptas, 2000; Arriaga and Vempala, 1999).

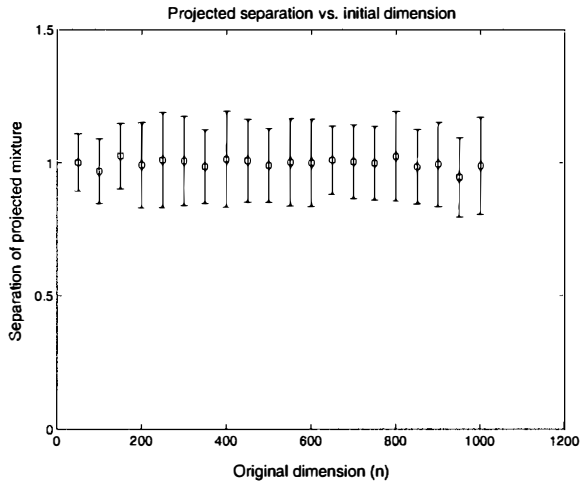

Figure 3: The projected dimension ($d = 20$) does not depend upon the original dimension.

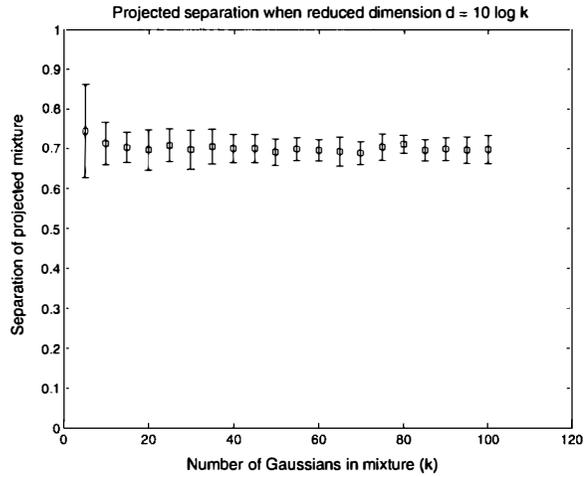

Figure 4: $d = 10 \ln k$ works nicely to keep the projected clusters well-separated.

### 3.3 Illustrative experiments

The lemmas of the previous section can be illustrated by a few simple experiments. The first two of these examine what happens when a 1-separated mixture of $k$ Gaussians in $\mathbb{R}^n$ is randomly projected into $\mathbb{R}^d$. The main question is, in order to achieve a fixed level of separation in the projected mixture, what value of $d$ must be chosen? How will this $d$ vary with $n$ and $k$?

The first experiment, depicted in Figure 3, is intended to demonstrate that $d$ does not depend upon $n$, that is, the projected dimension is independent of the original dimension of the data. Here two 1-separated spherical Gaussians are projected into $\mathbb{R}^{20}$ and their separation is noted as a function of $n$. The error bars are for one standard deviation in either direction; there are 40 trials per value of $n$.

The second series of tests (Figure 4) randomly projects 1-separated mixtures of $k$ spherical Gaussians in $\mathbb{R}^{100}$ into $d = 10 \ln k$ dimensions, and then notes the separation of the resulting mixtures. The results directly corroborate Lemma 1. The mixtures created for these tests are maximally packed, that is, each pair of constituent Gaussians is 1-separated. There are 40 measurements taken for each value of $k$.

The last two experiments, shown in Figures 5 and 6, document the dramatic decrease in eccentricity that can accompany the random projection of a Gaussian. The first of these projects a Gaussian of high eccentricity E from $\mathbb{R}^n$ into $\mathbb{R}^{20}$ and measures the eccentricity E$^*$ of the projection, over a range of values of E and $n$. Here matrices of eccentricity E are constructed by sampling the square roots of their eigenvalues uniformly from the range $[1, E]$, and making sure to include the endpoints 1 and E. The last experiment fixes a particular Gaussian in $\mathbb{R}^{50}$ of eccentricity

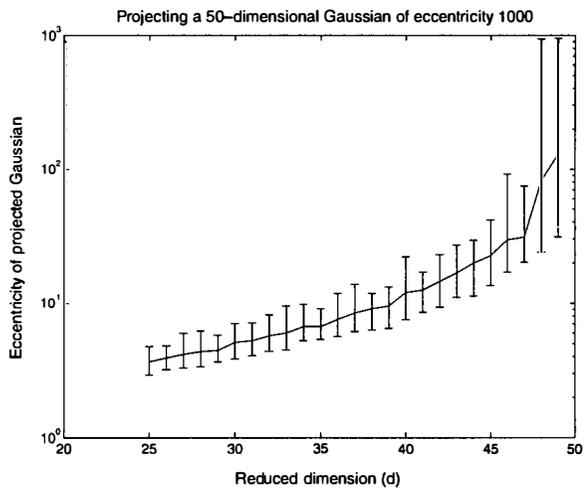

Figure 6: The eccentricity E$^*$ in $\mathbb{R}^d$ of a projected Gaussian whose original eccentricity in $\mathbb{R}^{50}$ is E $= 1,000$.

E $= 1,000$, and then projects this Gaussian into successively lower dimensions $49, 48, 47, \ldots, 25$. Notice that the $y$-axis of the graph has a logarithmic scale. Also, the error bars no longer represent standard deviations but instead span the maximum and minimum eccentricities observed over 40 trials per value of $d$.

We end this section with a simple experiment which will clarify some of the differences between random projection and PCA. Our earlier "bad example" (Section 3.1) was intended to show that PCA cannot in general reduce the dimension below $\Omega(k)$. Although it clearly demonstrated this, it required a symmetric arrangement of clusters that is unlikely to occur in practice. We feel that PCA suffers from a more fundamental flaw, namely, that it is apt to be confused by very eccentric Gaussians. This is not hard to visualize: in its attempt to pick important directions, it could easily be misled into choosing directions in which individual Gaussians have high variance, instead of directions of high intercluster distance.

| *Eccentricity* E | $n$ | | | | |
| in $\mathbb{R}^n$ | 25 | 50 | 75 | 100 | 200 |
|---|---|---|---|---|---|
| 50  | 9.5 ± 3.80  | 3.4 ± 0.62 | 2.5 ± 0.29 | 2.2 ± 0.17 | 1.7 ± 0.07 |
| 100 | 13.1 ± 5.79 | 3.5 ± 0.57 | 2.5 ± 0.26 | 2.2 ± 0.19 | 1.7 ± 0.08 |
| 150 | 13.0 ± 7.40 | 3.5 ± 0.55 | 2.5 ± 0.25 | 2.2 ± 0.14 | 1.7 ± 0.07 |
| 200 | 14.7 ± 8.04 | 3.4 ± 0.50 | 2.5 ± 0.22 | 2.2 ± 0.19 | 1.7 ± 0.06 |

Figure 5: Reduction in eccentricity E → E*, for a variety of starting dimensions $n$. Values in the table represent eccentricities E* in the projected space $\mathbb{R}^{20}$, plus or minus one standard deviation. Each table entry is the result of 40 trials.

In this experiment, data is generated from a 0.5-separated mixture of five Gaussians in $\mathbb{R}^{100}$. The Gaussians have different diagonal covariance matrices of eccentricity 1,000. The data are projected into 10-dimensional space using PCA and random projection, and the resulting separations, between all pairs of clusters, are shown in Figure 7. Random projection works predictably, preserving separation, but PCA virtually collapses all the clusters into one. It is fooled into picking directions which correspond to the eccentricity of individual Gaussians, and captures very little of the intercluster variation.

Random projection guarantees a certain level of performance regardless of the individual data set. PCA can occasionally do much better, but seems less reliable. Its usefulness varies from data set to data set, and it is particularly susceptible to problems in the case of highly eccentric clusters. Perhaps a hybrid scheme might work well, which first uses random projection to make the clusters more spherical and then applies PCA for a further reduction in dimension.

## 4 Random projection and EM

### 4.1 A simple algorithm

EM is at present the method of choice for learning mixtures of Gaussians and it is therefore important to find a way of integrating it with random projection. We suggest a very simple scheme.

- Project the data into a randomly chosen $d$-dimensional subspace.
- Run EM to convergence on the low-dimensional data.
- Take the fractional labels from EM's final soft-clustering of the low-dimensional data, and apply these same labels to the original high-dimensional data. This yields estimates of the high-dimensional means, covariances, and mixing weights.
- Run EM for one step in high dimension.

The final answer might not be a local maximum of the likelihood surface. If this is an issue, EM can be run for further rounds in the high-dimensional space; we have not done so in our experiments. We will also use a similar procedure, with PCA in place of random projection, for purposes of comparison.

### 4.2 Experiments on synthetic data

We have observed the performance of EM with and without random projection on data from a wide variety of Gaussian mixtures. We now discuss a few of these experiments.

The mixtures used were indexed by: the dimension, $n$; the number of Gaussians, $k$; the eccentricity of the covariance matrices, E; and the separation between the clusters, $c$. As in the previous section, a covariance matrix of eccentricity E was chosen by sampling the square roots of its eigenvalues uniformly from $[1, \text{E}]$, and making sure to always include the two endpoints 1, E. The Gaussians were positioned to be packed as tightly as possible subject to the $c$-separation requirement. Each mixing weight was chosen from a distribution which was not too far from uniform over $[\frac{1}{2k}, \frac{3}{2k}]$, and of course they summed to one. Let $w_i, \mu_i, \Sigma_i$ denote the true mixing weights, means, and covariance matrices, and let $w_i^{(t)}, \mu_i^{(t)}, \Sigma_i^{(t)}$ denote the estimated parameters at time $t$.

It is common to not allow each cluster its own full covariance matrix, because of the large number of parameters involved. We considered one typical restriction: where the Gaussians all share a single full covariance matrix.

There is also the important issue of initialization. We started all mixing weights $w_i^{(0)}$ equal, and chose the initial centers $\mu_i^{(0)}$ randomly from the data set. The corresponding covariance matrices were forced to be spherical at the outset, $\Sigma_i^{(0)} = \sigma_i^{(0)2} I$, with variance parameter

$$\sigma_i^{(0)2} = \frac{1}{2n} \min_{j \neq i} \|\mu_j^{(0)} - \mu_i^{(0)}\|^2.$$

We do not know the origin of this initializer but have found a slight variant of it mentioned in Bishop's (1995) text. If the clusters were restricted to have the same covariance, then the minimum of all the $\sigma_i^{(0)}$ was chosen as the common value.

|   | 1 | 2 | 3 | 4 | 5 |
|---|---|---|---|---|---|
| 1 | 0 | 0.04 | 0.03 | 0.03 | 0.02 |
| 2 | 0.04 | 0 | 0.03 | 0.04 | 0.03 |
| 3 | 0.03 | 0.03 | 0 | 0.03 | 0.03 |
| 4 | 0.03 | 0.04 | 0.03 | 0 | 0.03 |
| 5 | 0.02 | 0.03 | 0.03 | 0.03 | 0 |

|   | 1 | 2 | 3 | 4 | 5 |
|---|---|---|---|---|---|
| 1 | 0 | 0.57 | 0.60 | 0.44 | 0.41 |
| 2 | 0.57 | 0 | 0.68 | 0.43 | 0.43 |
| 3 | 0.60 | 0.68 | 0 | 0.64 | 0.45 |
| 4 | 0.44 | 0.43 | 0.64 | 0 | 0.37 |
| 5 | 0.41 | 0.43 | 0.45 | 0.37 | 0 |

Figure 7: A 0.5-separated mixture of five Gaussians in $\mathbb{R}^{100}$ is projected into $\mathbb{R}^{10}$ using PCA (on the left) and random projection (on the right). The resulting intercluster separations are shown.

For the first set of experiments we created mixtures of five Gaussians which were 1-separated and which had the same *spherical* covariance matrix ($\mathrm{E} = 1, k = 5, c = 1$). The training data consisted of 1,000 i.i.d. samples from such a mixture; an additional 1,000 samples were drawn as test data. We attempted to fit the training data with a mixture of five Gaussians with full common covariance matrix.

We used two measures of quality: (1) log-likelihood on the test data; (2) whether the true means were correctly learned. In the latter case, the final value of $\mu_i^{(t)}$ was considered a close enough approximation to $\mu_i$ if the $L_2$ error $\|\mu_i^{(t)} - \mu_i\|$ was at most a third the radius of the true Gaussian $N(\mu_i, \Sigma_i)$. Criterion (2) was satisfied only if *every* center was well-approximated in this way.

A variety of starting dimensions was chosen: $n = 50, 100, 150, 200$. For each, 40 mixtures were generated, and from each mixture, a training and test set were drawn. On each data set, we conducted 40 trials in which regular EM was matched against EM with random projection to $d = 25$ dimensions. In total therefore, 1,600 trials were conducted per value of $n$. The averaged results can be seen in Figure 8. The *success probability* denotes the fraction of trials in which *all* centers were correctly learned (in the approximate sense mentioned above). In these experiments, regular EM suffered a curse of dimensionality: as $n$ was increased, the success probability dropped sharply, and the number of iterations required for convergence increased gradually. In contrast, the performance of EM with random projection remained about stable. The last two rows of the table measure the percentage of trials in which our variant of EM was strictly better than regular EM (in terms of log-likelihood on the test data) and the percentage of trials on which the two were equally matched. Note that all likelihoods are with respect to the high-dimensional data and can therefore be compared. As the dimension $n$ rises, the advantage of using random projection with EM becomes more and more clear. The number of low-dimensional iterations needed by our adaptation of EM is slightly more than that required by regular EM; however, each of these low-dimensional iterations is much quicker, taking time proportional to $d^3$ instead of $n^3$.

Notice that the advantage conferred by random projection did not in this instance depend upon its eccentricity reduction properties, since the true clusters were spherical.

We also conducted the same tests using PCA in place of random projection. In terms of log-likelihoods on test data, both dimensionality reduction techniques performed about equally well.

In a second set of experiments, we generated data from a mixture of three 0.8-separated Gaussians with *different* covariance matrices of eccentricity 25 ($k = 3, \mathrm{E} = 25, c = 0.8$). We used a similar setup to that of the first experiment, with training and test sets consisting of 1,000 data points. This time, we allowed the learned Gaussians to have unrestricted and different covariances. There was again an advantage in using random projection. For instance, given data in $\mathbb{R}^{100}$, EM with projection down to 25 dimensions beat regular EM (in terms of log-likelihood on a test set) 69% of the time, and doubled the *success probability*, from 37% to 76%. At the same time, the average number of iterations required by EM on the projected data was 32.76, as opposed to 8.76 on the high-dimensional data. This increase was still more than offset by the time required per iteration: $100^3 = 25^3 \times 64$.

### 4.3 Experiments with OCR

To reassure ourselves that random projection is robust to the quirks of real data, we have used it as part of a simple classifier for handwritten digits. The training data consisted of 9,709 labeled instances of handwritten digits collected by USPS. Each was represented as a vector in $[-1, 1]^{256}$, corresponding to some $16 \times 16$ bitmap. A certain amount of preprocessing had been performed; details can be found in other studies which use this canonical data set, for instance that of Cortes and Vapnik (1995). There was a test set of 2,007 instances.

We constructed the classifier in a straightforward manner. First a random projection from $\mathbb{R}^{256}$ to $\mathbb{R}^d$ was fixed, for some small (double-digit) dimension $d$. Then the training data was projected into the low-dimensional space. A mixture of fifty Gaussians was fit to the data, five Gaussians per digit, by using EM on the instances of each digit separately. The five Gaussians for each digit had a shared covariance matrix; there were no other restrictions. No attempt was made to reconstruct the high-dimensional parameters.

| $n$ | 50 | 100 | 150 | 200 |
|---|---|---|---|---|
| Regular EM: | | | | |
| Success probability(%) | 43.5 | 36.6 | 29.2 | 23.1 |
| Average number of iterations | 14.81 | 14.19 | 15.10 | 16.62 |
| Random projection + EM: | | | | |
| Success probability(%) | 48.0 | 47.9 | 47.1 | 48.6 |
| Average number of iterations | 19.06 | 19.41 | 19.77 | 19.77 |
| Log-likelihood on test set: | | | | |
| Our EM exactly matched regular EM | 0.00 | 4.12 | 7.19 | 5.44 |
| Our EM beat regular EM | 52.81 | 54.94 | 60.25 | 66.94 |

Figure 8: A comparison of regular EM and our variant on test data of different dimensions $n$. The last two lines indicate the percentage of trials in which our method exactly matched or beat regular EM.

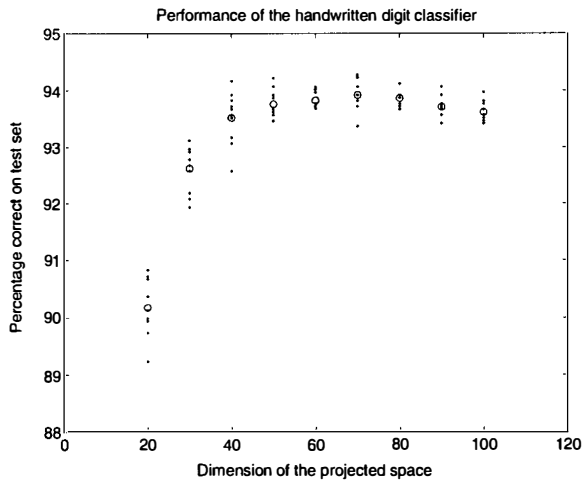

Figure 9: Performance of the digit classifier for a variety of projected dimensions. Dots indicate single trials, circles indicate averages.

In order to classify a test instance, it was first mapped into $d$-dimensional space using the projection which was fixed at the outset. Then the instance was assigned to the low-dimensional Gaussian which assigned it the highest conditional probability.

This process was repeated many times, for various values of $d$ ranging from 20 to 100. The results appear in Figure 9. The measure of performance is the usual one, percentage accuracy on the test set. It can be seen that the optimal performance of about 94% was attained at $d = 40$ or $d = 50$ and that higher values of $d$ did not noticeably improve performance. This corroborates our view throughout that even if data is high-dimensional, a random projection to a much smaller space can preserve enough information for clustering. The error rate was not exceptionally good (there are systems which have just over 4% error), but was reasonable for a classifier as naive as this.

We then performed a little bit of exploratory data analysis. In the original 256-dimensional space, we fit a Gaussian to (the instances of) each digit. We made a few measurements on the resulting ten Gaussians, to get an idea of the eccentricity of the individual clusters and the separation between them. The results are presented in Figure 10. The first point of interest is that the clusters have enormous eccentricity. It is easy to see why: for any given digit, some of the coordinates vary quite a bit (corresponding to central pixels) whereas other coordinates have negligible variance (for instance, border pixels). The resulting covariance matrices of the clusters are so ill-conditioned that it is difficult to reliably perform basic linear algebra operations, like determinant or inverse, on them. For this reason, we were unable to run EM in the original high-dimensional space. Various possible fixes include adding Gaussian noise to the data or placing strong restrictions on EM's covariance estimates. The key point is that using random projection, we were able to run EM on the data with no tweaking whatsoever.

Why exactly was this possible? To find the answer, we picked one random projection into 40-dimensional space, and carried out the same analysis as above on the projected data. The results can be seen in Figure 11. The separation between the clusters remained about the same as in the original space, and the eccentricity of each cluster was reduced drastically, corroborating the theoretical results presented earlier.

We also conducted these experiments using PCA, and once again, the optimal success rate of about 94% was obtained using a projection to $d \geq 40$ dimensions.

## 5 An open problem

This paper attempts to document carefully some of the benefits of using random projection for learning mixtures of Gaussians. But what about other mixture models? For instance, mixtures of tree-structured distributions have been used effectively in the recent machine learning literature, by Meilă, Jordan, and Morris (1998).

A fascinating result of Diaconis and Freedman (1984)

| Digit | 0 | 1 | 2 | 3 | 4 | 5 | 6 | 7 | 8 | 9 | Eccentricity |
|---|---|---|---|---|---|---|---|---|---|---|---|
| 0 | 0.00 | 1.61 | 0.93 | 1.01 | 1.18 | 0.88 | 0.82 | 1.31 | 1.00 | 1.11 | $1.32 \times 10^8$ |
| 1 | 1.61 | 0.00 | 1.04 | 1.38 | 1.06 | 1.27 | 1.29 | 1.19 | 1.23 | 1.22 | $4.78 \times 10^{13}$ |
| 2 | 0.93 | 1.04 | 0.00 | 0.70 | 0.81 | 0.82 | 0.78 | 0.75 | 0.63 | 0.78 | $1.13 \times 10^8$ |
| 3 | 1.01 | 1.38 | 0.70 | 0.00 | 0.97 | 0.65 | 1.02 | 0.98 | 0.71 | 0.88 | $4.86 \times 10^5$ |
| 4 | 1.18 | 1.06 | 0.81 | 0.97 | 0.00 | 0.97 | 0.90 | 0.91 | 0.77 | 0.64 | $4.22 \times 10^8$ |
| 5 | 0.88 | 1.27 | 0.82 | 0.65 | 0.97 | 0.00 | 0.70 | 1.06 | 0.70 | 0.85 | $1.88 \times 10^4$ |
| 6 | 0.82 | 1.29 | 0.78 | 1.02 | 0.90 | 0.70 | 0.00 | 1.31 | 0.83 | 1.09 | $6.84 \times 10^8$ |
| 7 | 1.31 | 1.19 | 0.75 | 0.98 | 0.91 | 1.06 | 1.31 | 0.00 | 0.97 | 0.70 | $6.02 \times 10^{11}$ |
| 8 | 1.00 | 1.23 | 0.63 | 0.71 | 0.77 | 0.70 | 0.83 | 0.97 | 0.00 | 0.65 | $2.32 \times 10^8$ |
| 9 | 1.11 | 1.22 | 0.78 | 0.88 | 0.64 | 0.85 | 1.09 | 0.70 | 0.65 | 0.00 | $1.04 \times 10^9$ |

Figure 10: Separation between the digits in $\mathbb{R}^{256}$, that is, before projection. The rightmost column indicates the eccentricity of each digit's cluster.

| Digit | 0 | 1 | 2 | 3 | 4 | 5 | 6 | 7 | 8 | 9 | Eccentricity |
|---|---|---|---|---|---|---|---|---|---|---|---|
| 0 | 0.00 | 1.58 | 0.96 | 1.01 | 1.14 | 0.81 | 0.83 | 1.23 | 1.07 | 1.10 | 31.51 |
| 1 | 1.58 | 0.00 | 1.07 | 1.23 | 1.03 | 1.30 | 1.45 | 1.20 | 1.24 | 1.17 | 66.41 |
| 2 | 0.96 | 1.07 | 0.00 | 0.73 | 0.76 | 0.88 | 0.82 | 0.68 | 0.58 | 0.63 | 24.25 |
| 3 | 1.01 | 1.23 | 0.73 | 0.00 | 0.97 | 0.77 | 1.13 | 0.87 | 0.76 | 0.80 | 17.87 |
| 4 | 1.14 | 1.03 | 0.76 | 0.97 | 0.00 | 0.87 | 0.79 | 0.84 | 0.72 | 0.57 | 28.06 |
| 5 | 0.81 | 1.30 | 0.88 | 0.77 | 0.87 | 0.00 | 0.71 | 1.08 | 0.84 | 0.81 | 18.72 |
| 6 | 0.83 | 1.45 | 0.82 | 1.13 | 0.79 | 0.71 | 0.00 | 1.34 | 0.91 | 1.06 | 25.53 |
| 7 | 1.23 | 1.20 | 0.68 | 0.87 | 0.84 | 1.08 | 1.34 | 0.00 | 0.87 | 0.61 | 34.01 |
| 8 | 1.07 | 1.24 | 0.58 | 0.76 | 0.72 | 0.84 | 0.91 | 0.87 | 0.00 | 0.50 | 23.37 |
| 9 | 1.10 | 1.17 | 0.63 | 0.80 | 0.57 | 0.81 | 1.06 | 0.61 | 0.50 | 0.00 | 32.21 |

Figure 11: Separation between the digits in $\mathbb{R}^{40}$, that is, after projection, and the eccentricity of each cluster.

shows that a wide variety of high-dimensional distributions look more Gaussian when randomly projected into a low-dimensional subspace. This follows from central limit theorems, specifically the Berry-Esséen theorem (Feller, 1966), and suggests an unusual methodology for learning a broad class of mixture models: randomly project the data, fit a mixture of Gaussians to the resulting low-dimensional data to cluster it, and then use the induced (soft or hard) clustering in the original high-dimensional space to learn the high-dimensional component distributions. This promising scheme merits much closer study.